\documentclass[runningheads]{llncs}
\usepackage[T1]{fontenc}
\usepackage{graphicx,verbatim}
\usepackage{amsmath}
\usepackage{amssymb}
\usepackage[colorlinks, linkcolor=blue, citecolor=blue, urlcolor=blue]{hyperref}

\begin{document}
\bibliographystyle{unsrt}  





\title{NeuroDx-LM: A Clinical Large-Scale Model for EEG-based Neurological Disorder Detection}
\titlerunning{NeuroDx-LM: A Clinical Large-Scale Model}

\author{
  Guanghao Jin\inst{1,2,*} \and
  Yuan Liang\inst{1,*} \and
  Yihan Ma\inst{1} \and
  Jingpei Wu\inst{2} \and
  Guoyang Liu\inst{1,\dagger}
}
\authorrunning{Guanghao Jin et al.}

\institute{
  Shandong University, China \and LMU Munich, Germany
   \\
  \email{jin@cip.ifi.lmu.de, 202335720@mail.sdu.edu.cn, 202336133@mail.sdu.edu.cn,} \\
  \email{jingpei.wu@lmu.de, gyliu@sdu.edu.cn}
}

\maketitle

\begin{abstract}
Large-scale models pre-trained on Electroencephalography (EEG) have shown promise in clinical applications such as neurological disorder detection. However, the practical deployment of EEG-based large-scale models faces critical challenges such as limited labeled EEG data and suboptimal performance in clinical scenarios. To address these issues, we propose NeuroDx-LM, a novel large-scale model specifically designed for detecting EEG-based neurological disorders. Our key contributions include (i) a Selective Temporal-Frequency Embedding mechanism that adaptively captures complex temporal and spectral patterns in EEG signals; and (ii) a Progressive Feature-Aware Training strategy that refines feature representation in a two-stage process. In the first stage, our model learns the fundamental discriminative features of EEG activities; in the second stage, the model further extracts more specialized fine-grained features for accurate diagnostic performance. We evaluated NeuroDx-LM on the CHB-MIT and Schizophrenia datasets, achieving state-of-the-art performance in EEG-based seizure and schizophrenia detection, respectively. These results demonstrate the great potential of EEG-based large-scale models to advance clinical applicability. Our code is available at \url{https://github.com/LetItBe12345/NeuroDx-LM}.


\end{abstract}
\section{Introduction}

Neurological disorders, including epilepsy and schizophrenia, affect over 3 billion patients worldwide, causing an annual economic loss of around 2.5 trillion dollars\cite{steinmetz2024global,li2025health}. Electroencephalography (EEG) as a method for recording the electrical activity of the brain plays a critical role in detecting neurological disorders. Given that the traditional EEG analysis is conducted by well-trained neurologists, which is time-consuming, automatic deep learning-based approaches are highly demanded.

To date, various deep learning methods have been proposed for EEG analysis \cite{singh2023early,iqbal2024progress}. For instance, Ullah \textit{et al.}~\cite{ullah2018automated} adopted 2D convolutional architectures to capture local spatiotemporal patterns from EEG signals for seizure detection. Lee \textit{et al.}~\cite{lee2019deep} further incorporated temporal dependency modeling through recurrent components. However, these models often have limited generalization ability, making it challenging to capture robust EEG features fitting to different conditions and noises. 

Recent breakthroughs in large language models (LLMs)\cite{liu2024survey,ouyang2022training} have introduced novel frameworks for unified EEG representation learning. Multiple studies attempted to address these challenges through masked EEG pre-training frameworks enabling models to develop a universal understanding of EEG data for rapid adaptation to various downstream tasks. For example, LaBraM\cite{jiang2024large} introduced a neural tokenizer that employs masked neural code prediction to establish foundational representations of brain activities. Wang \textit{et al.}~\cite{wang2025eegpt} proposed EEGPT, a novel framework leveraging LaBraM and introducing a dual self-supervised EEG universal representation learning method based on the spatio-temporal consistency of EEG signals. This approach enables efficient linear-probing adaptation, achieving state-of-the-art performance across multiple EEG analysis tasks.

Although large-scale model pre-trained on EEG data have partially alleviated the aforementioned challenges, their clinical applicability remains limited. Specifically, their effectiveness is hindered by several factors, including the low signal-to-noise ratio (SNR) inherent in EEG signals, the complex and multi-faceted nature of neurological disorders, and the limited availability of annotated EEG datasets. These challenges obstruct the transferability of EEG feature representations learned through pre-training to downstream tasks, resulting in suboptimal performance in tasks such as neurological disorder detection. 
Furthermore, conventional pre-training approaches predominantly rely on full-parameter fine-tuning, which imposes high GPU memory demands during training, significantly increasing the cost of deployment.

To further address the aforementioned challenges, this paper proposes the NeuroDx-LM, a large-scale model with more generalization ability and interpretability, which inspired by specialized medical training programs and curriculum learning\cite{bengio2009curriculum,liu2024let} for EEG-based disease diagnosis. The main contributions of this work are as follows:  


1) A Selective Temporal-Frequency Embedding Module is proposed to adaptively fuse frequency-domain features, obtained by applying convolution to EEG signals after the Fast Fourier Transform (FFT), into time-domain features, addressing the commonly overlooked frequency patterns of EEG signals in large-scale EEG models.

2) A two-stage progressive feature-aware training strategy is introduced, which sequentially enhances the capability of large-scale model in EEG-based disease diagnosis:  
   i) Leveraging large-scale EEG data labeled as normal/abnormal with temporal embeddings to capture more fundamental and discriminative EEG features.  
   ii) Fine-tuning the model on smaller, high-quality disease-specific datasets with frequency-domain features to improve the performance of EEG-based disease detection.

3) Extensive experiments on two EEG datasets demonstrate that NeuroDx-LM achieves state-of-the-art performance across multiple evaluation metrics in EEG-based disease detection while reducing GPU memory usage during training.

\section{Materials and Methods}

\subsection{Datasets}
\textbf{TUAB Dataset.} This dataset was collected from Temple University, and it consists of 23-channel EEG recordings sampled at 256 Hz, totaling 1169 hours of EEG data. All recordings have been annotated as normal or abnormal \cite{obeid2016temple}. This dataset was only used for the first stage of model training. 

\noindent\textbf{CHB-MIT Dataset.} This dataset includes scalp EEG recordings obtained from 22 pediatric epilepsy patients at Children’s Hospital Boston. It comprises a total of 961 hours of EEG data with 198 recorded seizure events. All EEG signals were acquired at a sampling rate of 256 Hz and the seizure onsets were labeled by experienced clinical neurologists.\cite{shoeb2009application,guttag2010chb}.

\noindent\textbf{Schizophrenia Dataset.} This dataset includes scalp EEG recordings from 14 schizophrenia patients and 14 healthy control patients, collected as part of a study conducted at the Institute of Psychiatry and Neurology in Warsaw, Poland. EEG signals were acquired using 19 electrodes positioned according to the International 10-20 system, with a sampling rate of 250 Hz.\cite{repod.0107441_2017}

\subsection{Preprocessing}

The preprocessing of EEG signals is essential for improving their quality which in turn improves their applicability in the clinical diagnosis of neurological disorders. In this study, we implemented a standardized preprocessing pipeline across all three EEG datasets. First, channels that did not conform to the International 10-20 system were excluded. Next, a bandpass filter (0.1–75 Hz) was applied to remove low-frequency drift and high-frequency noise. To mitigate power-line interference, a 50 Hz notch filter was employed. Additionally, all EEG signals were resampled to 200 Hz to ensure uniform sampling across datasets. Finally, before being fed into the model, the EEG data from each dataset were segmented into non-overlapping 10-second segments to facilitate subsequent analysis.

\begin{figure}[htbp]
    \centering
    \includegraphics[width=\textwidth]{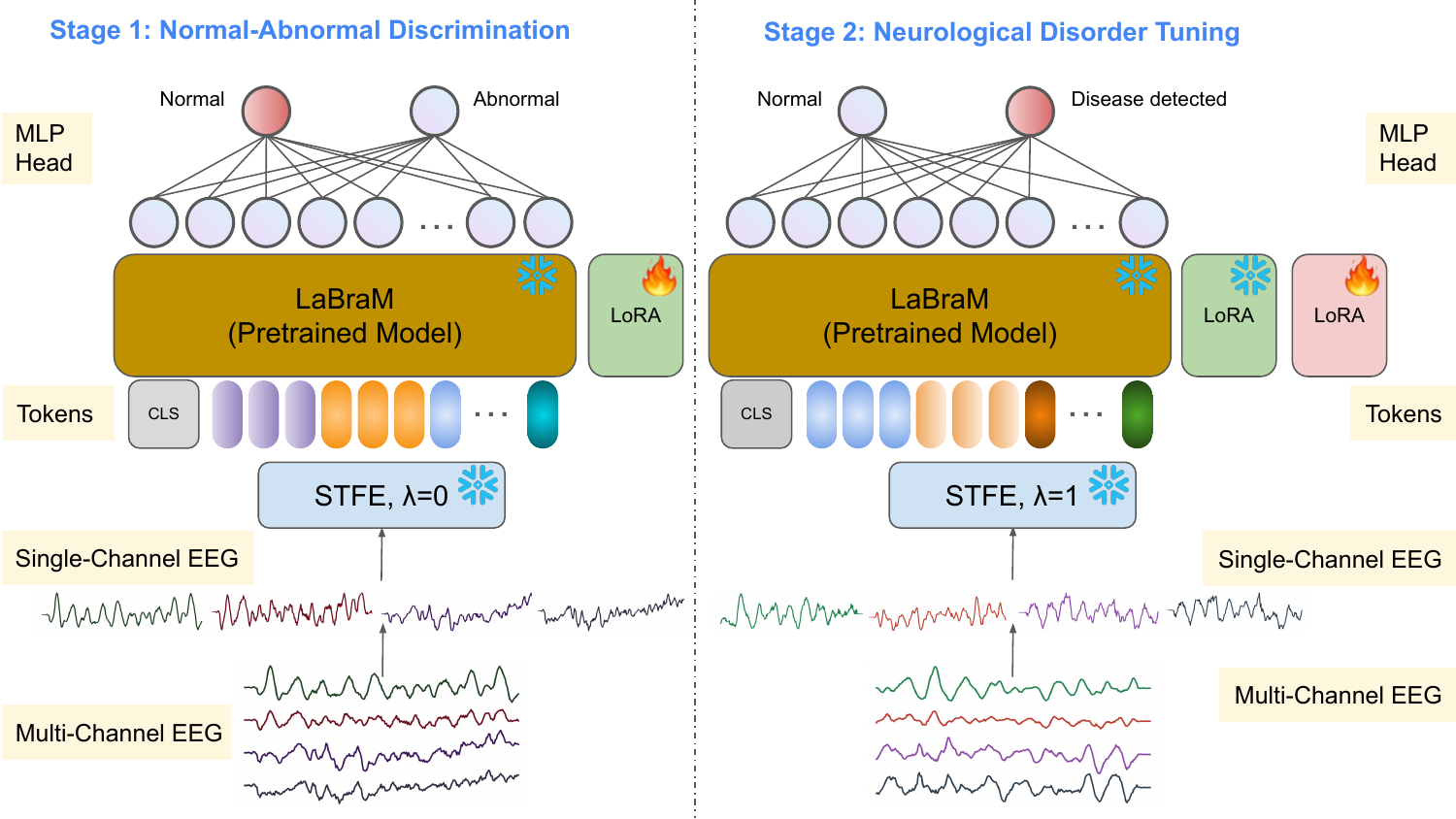}
    \caption{Our two-stage progressive feature-aware training framework. In the first stage, we train NeuroDx-LM on TUAB dataset to possess robust discrimination between normal and abnormal EEG signals. In the second stage, we use
 high-quality EEG dataset related to a specific neurological disorder for fine-tuning, which enables more precise neurological disorder detection.}
    \label{fig:model_architecture}
\end{figure}

\subsection{Architecture of the NeuroDx-LM Model }
In this section, we introduce the NeuroDx-LM Model architecture, as shown in Figure~\ref{fig:model_architecture}. Before being fed into the model, each preprocessed EEG signal is segmented into 10 non-overlapping 1-second patches. Each EEG patch is denoted as \( x_{i,j} \in \mathbb{R}^{C \times A \times T} \) where \( C \) represents the number of channels, \( A \) refers to the number of patches, and \( T \) stands for the time steps.

\begin{figure}[htbp]
    \centering
    \includegraphics[width=\textwidth]{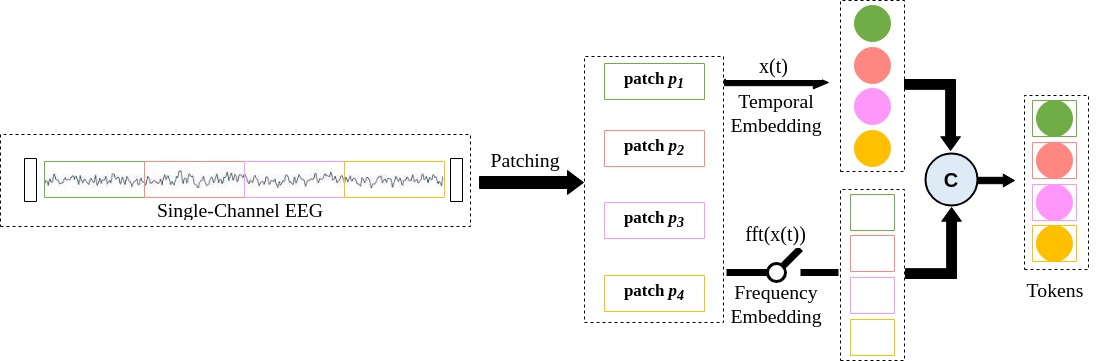}
    \caption{An overview of Selective Temporal-Frequency Embedding module.}
    \label{fig:STFE}
\end{figure}

\subsubsection{Selective Temporal-Frequency Embedding (STFE)}
To capture both the abnormal temporal synchronization and the distinct rhythmic (frequency-related) abnormalities in EEG signals, we propose the STFE module (see Figure~\ref{fig:STFE}). It consists of two main branches—temporal embedding and frequency embedding—and a mechanism to selectively fuse them.

\begin{itemize}
    \item \textbf{Temporal Embedding (TE):} We employ a multi-scale temporal analysis method to capture EEG features at different temporal scales. Three independent temporal convolutional blocks were designed, each consisting of a convolution\cite{lecun1995convolutional} in different domains, a GELU\cite{hendrycks2016gaussian}, and Group Normalization\cite{wu2018group}. Each block can be expressed as:

\begin{equation}
z^{\text{time}'} = \text{GN}(\text{GELU}(\text{Conv}(x(t))))
\end{equation}

where \(\text x(t) \) denotes input EEG signals, \( \text{Conv} \) represents a temporal convolution operation, and GN denotes Group Normalization. The sizes of these three convolutional kernels are set to \(\{15, 3, 3\}\), respectively.

    \item \textbf{Frequency Embedding (FE) with Fast Fourier Transform:} 
    We perform a Fast Fourier Transform (FFT) on the time-domain EEG signal to obtain its frequency-domain representation. After performing FFT, we retain the amplitude spectrum and apply four convolutional blocks of varying kernel sizes. 
    The sizes of these four convolutional kernels are set to \{3, 15, 3, 3\}, respectively.  
    Each block can be expressed:  
\begin{equation}
    z^{\text{freq}^{\prime}} =
    \operatorname{GN} \left( 
        \operatorname{GELU} \left( 
            \operatorname{Conv} \left( 
                \left| \mathcal{F}\{x(t)\} \right| 
            \right) 
        \right) 
    \right) 
\end{equation}

    \noindent where $\mathcal{F}\{\cdot\}$ represents the Fourier transform implemented by FFT.  This method can capture discriminative frequency features in terms of energy distribution, which is often linked to epileptiform or psychotic episodes.
    
    \item \textbf{Selective Fusion:} In order to accommodate the varying feature requirements at different stages of training, we devise a dynamic gating variable \( \lambda_f \in [0,1] \) that adaptively modulates the contribution of frequency-domain features throughout the training process.The final representation of the \( i \)-th channel in the \( j \)-th patch is given by:
        \begin{equation}
        z_{i,j} = W \left[ z^{\text{time}}_{i,j} \mathbin{\textcircled{c}} (\lambda_f \cdot z^{\text{freq}}_{i,j}) \right]
        \end{equation}

    where \( W \) is a linear transformation and $\textcircled{c}$ denotes concatenating operation.
\end{itemize}

\subsubsection{Input of Large-Scale Model.} 

For each patch, we map it into an \( R^d \) vector space using the Selective Temporal-Frequency Embedding layer, where \( d \) represents the hidden dimension of the large-scale model pre-trained on EEG . This process results in an embedded sequence \(\{z_1, z_2, \dots, z_M\}\), where \( z_i \in R^d \), and \( M \) denotes the number of patches. To adapt this input for a pre-trained large-scale model (e.g., LaBraM), we prepend a special classification token (\(\langle cls \rangle\)), forming the final input sequence $[\langle cls \rangle, z_1, z_2, \dots, z_M]$. This sequence is then fed into the pre-trained large-scale model for further encoding, enabling the extraction of high-level representations for subsequent classification tasks.

\subsection{Progressive Feature-aware Training}
The large-scale model pre-trained using masked EEG modeling requires adaptation to downstream tasks. In contrast to the previous typical one-stage training approaches, inspired by curriculum learning\cite{bengio2009curriculum,liu2024let}, our method proposes a two-stage training method to progressively enhance the model's ability to extract EEG signal features related to neurological disorders.

\subsubsection{Stage 1: Normal-Abnormal Discrimination. }
In Stage 1, we train the model on the TUAB dataset with long-duration EEGs labeled as normal/abnormal.
We freeze both the STFE module and the main model parameters, inserting a LoRA module\cite{hu2022lora} to handle the adaptation.
Specifically, for a pre-trained weight matrix $W_0 \in \mathbb{R}^{d \times k}$, we introduce a low-rank 
decomposition $\Delta W = BA$ with $B \in \mathbb{R}^{d \times r}$ and $A \in \mathbb{R}^{r \times k}$, where $r \ll \min(d,k)$. 
Accordingly, the forward pass becomes
\begin{equation}
\label{eq:LoRA-forward}
h = (W_0 + \Delta W)x = W_0 x + BAx = W_0 x + {BAx}
\end{equation}
where $x \in \mathbb{R}^{k \times 1}$ and $h \in \mathbb{R}^{d \times 1}$. During Stage~1, 
only $B$ and $A$ are updated, while $W_0$ is kept frozen. We set $\lambda_f = 0$ within the 
STFE module, thus disabling its frequency-domain branch and relying solely on time-domain 
features to capture obvious pathological waveforms (e.g., spikes, sharp waves). This design 
allows the model to learn a robust discrimination between normal and abnormal EEG signals without 
prematurely introducing frequency-domain complexity, which could otherwise lead to overfitting or unstable training at this early stage.

\subsubsection{Stage 2: Neurological Disorder Tuning.}
Next, we perform fine-tuning on disease-specific EEG dataset corresponding to each target task (e.g., epilepsy or schizophrenia detection). 
Due to patient privacy protection and the high cost of annotation, this dataset is usually limited in scale.
We now enable $\lambda_f = 1$ within the STFE module to incorporate the frequency-domain branch, which is essential for detecting periodic or rhythmic abnormalities, as specific neurological disorders exhibit more significant signal representations in the frequency domain compared to the time domain.
We then merge the LoRA module from Stage~1 into the base model, and introduce a new LoRA module $(B^{\text{disease}}, A^{\text{disease}})$. 
Thus, the updated weight matrix for Stage~2 is expressed as:
\begin{equation}
\tilde{W}_0 = W_0 + B^{\text{stage-1}} A^{\text{stage-1}}
\end{equation}

\noindent Consequently, the forward pass can be written as $h'=\tilde{W}_0x+B^{\text{disease}} A^{\text{disease}}x$, where $\tilde{W}_0$ is kept frozen, which ensures that the general abnormality-detection capability from Stage 1 is effectively retained and transferred, allowing more nuanced disease-specific detection in Stage 2.

\section{Experiments}
\subsection{Experimental Setup}
In this study, we adopt LaBraM-Base as the pre-trained large-scale model. For the CHB-MIT and Schizophrenia datasets, we follow the same data splitting strategy as LaBraM\cite{jiang2024large} for the TUAB dataset, dividing the data into training, validation, and test sets. During training, we use a batch size of 64. The optimizer is AdamW, incorporating layer decay and a warm-up strategy. In the first training stage, the model is trained for 1 epoch with a learning rate of \( 5 \times 10^{-4} \). In the second stage, the model is trained for 20 epochs with a learning rate of \( 1 \times 10^{-4} \). The LoRA parameters are set as \( r = 8 \), \( \alpha = 32 \). 
We adopt Accuracy, ROC-AUC, and PR-AUC as classification metrics.

\subsection{Comparison with state-of-the-art methods}
Table~\ref{tab:comparison} presents the performance of our proposed NeuroDx-LM model on the CHB-MIT and Schizophrenia datasets, compared to state-of-the-art baseline models. The experimental results indicate that NeuroDx-LM not only outperforms baselines on the majority of evaluation metrics for the CHB-MIT dataset, but also achieves significant improvements across all metrics in the Schizophrenia dataset. This highlights that the Progressive Feature-aware Training strategy enables our model to perform even better in few-shot learning scenarios.
\begin{table}[]
\centering
\caption{The results of different methods on CHB-MIT and Schizophrenia}
\label{tab:comparison}
\resizebox{\textwidth}{!}{
\begin{tabular}{lccccccc}
\hline
\textbf{}                          &            & \multicolumn{3}{c}{CHB-MIT}                         & \multicolumn{3}{c}{Schizophrenia}                   \\ \hline
Methods                            & Model Size & Accuracy        & ROC-AUC         & PR-AUC          & Accuracy        & ROC-AUC         & PR-AUC          \\ \hline
SPaRCNet\cite{jing2023development}        & 0.79M      & 0.7231          & 0.5814          & 0.2509          & 0.7231          & 0.5814          & 0.2509          \\
CNN-Transformer\cite{peh2022transformer} & 3.2M       & 0.7288          & 0.5685          & 0.2573          & 0.4513          & 0.4202          & 0.4677          \\
ST-Transformer\cite{song2021transformer} & 3.5M       & 0.7418          & 0.5175          & 0.2178          & 0.6580          & 0.7691          & 0.7735          \\
BIOT\cite{yang2023biot}          & 3.2M       & 0.7244          & \textbf{0.6185} & 0.2609          & 0.5781          & 0.5765          & 0.5719          \\
EEGPT\cite{wang2025eegpt}                              & 4.7M       & 0.7902          & 0.5377          & 0.2213          & 0.5139          & 0.5186          & 0.5534          \\
LaBraM\cite{wu2018group}                             & 5.8M       & 0.7550          & 0.5796          & 0.2447          & 0.7587          & 0.8668          & 0.8991          \\ \hline
NeuroDx-LM                         & 5.8M       & \textbf{0.7903} & 0.5903          & \textbf{0.2617} & \textbf{0.8246} & \textbf{0.8965} & \textbf{0.9114} \\ \hline
\end{tabular}
}
\end{table}

\subsection{Ablation Study}
In this section, we present a detailed ablation study of NeuroDx-LM in Table~\ref{tab:Ablation}. 
Our experimental results demonstrate that the proposed model (see Experiment 1) achieves the best performance in most metrics across both datasets. 
The key questions and corresponding answers in our ablation study are provided below.
\subsubsection{Q1: Why use Selective Temporal-Frequency Embedding?}
By comparing Experiment 1, Experiment 3, and Experiment 7, we observe that employing STFE yields superior results compared to using TE or TE+FE in both Stage 1 and Stage 2. This advantage is particularly significant in the Schizophrenia dataset, indicating that STFE facilitates better transfer of the normal/abnormal EEG Discrimination ability learned in Stage 1 to the detection of specific neurological diseases.
\subsubsection{Q2: Is every training stage necessary?}
By comparing Experiment 1, 5, and 9, we can observe a significant performance gap in the classification task when skipping Stage 1 training. On the CHB-MIT dataset, the model performs relatively well, whereas on the Schizophrenia dataset, the scores drop abnormally low. Further analysis of dataset scales reveals that CHB-MIT is 100 times larger than the Schizophrenia dataset. This suggests that for datasets with very limited scale, training only in Stage 2 makes it difficult for the model to establish a robust disease detection capability. In contrast, models trained with Stage 1 consistently outperform those trained directly in Stage 2 across all evaluation metrics.
\subsubsection{Q3: What are the advantages of using LoRA?}
By using LoRA instead of full-parameter fine-tuning, GPU memory usage is reduced by 7\%. Our experiments are performed on CUDA 11.8 with an RTX 3090 GPU, with the batch size set to 64 for model training. 
\begin{table}[!tb]
\centering
\caption{Ablation study on the two-stage training strategy. In Embedding, ''STFE'' (Selective Temporal-Frequency Embedding), ''TE'' (Temporal Embedding), and ''FE'' (Frequency Embedding with Fast Fourier Transform) denote different EEG feature extraction methods. In Stage 1, ''Yes'' or ''No'' indicates whether training is performed. In Stage 2, ''Reuse'' means reusing LoRA from Stage 1, ''Addition'' refers to adding a new LoRA module, while ''Full Parameter'' means using full-parameter fine-tuning.}
\label{tab:Ablation}
\resizebox{\textwidth}{!}{
\begin{tabular}{|cccc|ccc|ccc|}
\hline
\multicolumn{4}{|c|}{Configurations}                                                                                    & \multicolumn{3}{c|}{CHB-MIT}                                                                  & \multicolumn{3}{c|}{Schizophrenia}                                                            \\ \hline
\multicolumn{1}{|c|}{No.} & \multicolumn{1}{c|}{Embedding} & \multicolumn{1}{c|}{Stage1} & Stage2         & \multicolumn{1}{c|}{Accuracy}        & \multicolumn{1}{c|}{ROC-AUC}        & PR-AUC         & \multicolumn{1}{c|}{Accuracy}        & \multicolumn{1}{c|}{ROC-AUC}        & PR-AUC         \\ \hline
\multicolumn{1}{|c|}{1}   & \multicolumn{1}{c|}{STFE}      & \multicolumn{1}{c|}{Yes}   & Addition       & \multicolumn{1}{c|}{\textbf{0.7903}} & \multicolumn{1}{c|}{0.5903}          & 0.2617          & \multicolumn{1}{c|}{\textbf{0.8247}} & \multicolumn{1}{c|}{\textbf{0.8965}} & \textbf{0.9114} \\ \cline{1-1} \cline{4-4}
\multicolumn{1}{|c|}{2}   & \multicolumn{1}{c|}{}          & \multicolumn{1}{c|}{}       & Reuse          & \multicolumn{1}{c|}{0.7850}          & \multicolumn{1}{c|}{0.5557}          & 0.2314          & \multicolumn{1}{c|}{0.7569}          & \multicolumn{1}{c|}{0.8534}          & 0.8556          \\ \cline{1-4}
\multicolumn{1}{|c|}{3}   & \multicolumn{1}{c|}{}          & \multicolumn{1}{c|}{Yes}   & Addition       & \multicolumn{1}{c|}{0.7870}          & \multicolumn{1}{c|}{0.5738}          & 0.2445          & \multicolumn{1}{c|}{0.6541}          & \multicolumn{1}{c|}{0.6756}          & 0.7424          \\ \cline{1-1} \cline{4-4}
\multicolumn{1}{|c|}{4}   & \multicolumn{1}{c|}{TE+FE}       & \multicolumn{1}{c|}{}       & Reuse          & \multicolumn{1}{c|}{0.7627}          & \multicolumn{1}{c|}{0.5439}          & 0.2218          & \multicolumn{1}{c|}{0.7170}          & \multicolumn{1}{c|}{0.7438}          & 0.7778          \\ \cline{1-1} \cline{3-4}
\multicolumn{1}{|c|}{5}   & \multicolumn{1}{c|}{}          & \multicolumn{1}{c|}{No}     & Addition       & \multicolumn{1}{c|}{0.7902}          & \multicolumn{1}{c|}{0.5241}          & 0.2165          & \multicolumn{1}{c|}{0.7344}          & \multicolumn{1}{c|}{0.8536}          & 0.8805          \\ \cline{1-1} \cline{4-4}
\multicolumn{1}{|c|}{6}   & \multicolumn{1}{c|}{}          & \multicolumn{1}{c|}{}       & Full Parameter & \multicolumn{1}{c|}{0.7881}          & \multicolumn{1}{c|}{0.5763}          & 0.2149          & \multicolumn{1}{c|}{0.7813}          & \multicolumn{1}{c|}{0.8564}          & 0.8557          \\ \cline{1-4}
\multicolumn{1}{|c|}{7}   & \multicolumn{1}{c|}{}          & \multicolumn{1}{c|}{Yes}   & Addition       & \multicolumn{1}{c|}{0.7526}          & \multicolumn{1}{c|}{0.6027}          & 0.2501          & \multicolumn{1}{c|}{0.6285}          & \multicolumn{1}{c|}{0.7485}          & 0.7651          \\ \cline{1-1} \cline{4-4}
\multicolumn{1}{|c|}{8}   & \multicolumn{1}{c|}{TE}        & \multicolumn{1}{c|}{}       & Reuse          & \multicolumn{1}{c|}{0.7481}          & \multicolumn{1}{c|}{0.5997}          & 0.2474          & \multicolumn{1}{c|}{0.6944}          & \multicolumn{1}{c|}{0.7957}          & 0.7902          \\ \cline{1-1} \cline{3-4}
\multicolumn{1}{|c|}{9}   & \multicolumn{1}{c|}{}          & \multicolumn{1}{c|}{No}     & Addition       & \multicolumn{1}{c|}{0.7476}          & \multicolumn{1}{c|}{\textbf{0.6126}} & \textbf{0.2631} & \multicolumn{1}{c|}{0.7396}          & \multicolumn{1}{c|}{0.8541}          & 0.8804          \\ \cline{1-1} \cline{4-4}
\multicolumn{1}{|c|}{10}  & \multicolumn{1}{c|}{}          & \multicolumn{1}{c|}{}       & Full Parameter & \multicolumn{1}{c|}{0.7550}          & \multicolumn{1}{c|}{0.5796}          & 0.2447          & \multicolumn{1}{c|}{0.7274}          & \multicolumn{1}{c|}{0.8293}          & 0.8256          \\ \hline
\end{tabular}
}
\end{table}
\section{Conclusion}
In this work, we propose NeuroDx-LM, a model for the robust and accurate detection of neurological disorders. To enhance the model generalization ability, we introduce the Selective Temporal-Frequency Embedding Module that can extract robust EEG features from both time and frequency domains. Additionally, we propose a two-stage Progressive Feature-Aware Training strategy, which further enhances the performance and of the NeuroDx-LM by knowledge transferring, particularly in few-shot learning scenarios. Experimental results demonstrate that our approach achieves state-of-the-art performance on the CHB-MIT and Schizophrenia datasets. Furthermore, extensive ablation studies suggest the great robustness of the proposed method. 
\bibliography{refs}

\end{document}